# The Multi-fingered Kinematic Model for Dual-arm Manipulation

**Jingyi Li*[1]**

Polytechnic University of Catalonia, ETSEIB, SPAIN.
(E-mail: jingyi.li1@upc.edu)

**Highlights**

1. A planar kinematic model in the hand-object coordinates system for bimanual manipulation is presented. It can compute and determine the fingers configurations.
2. In the hand-object plane, the kinematic solutions from the model can generate valid manipulation strategies for the dual-arm robots.
3. In our experiment, the desired positions, as the model inputs are successfully generated valid joints values for bimanual manipulation.

**Abstract**

This paper presents the planar finger kinematic model for dual-arm robot to determine manipulation strategies. The first step is to model based on planar geometric features of the coordinated and rolling motion so that the robot can select the fingers configurations. For the hand-object model, we consider the distances between object and hands as the constraints. The second step is to seek the appropriate values of finger joints based on their positions samples which are randomly generated. Here the robot selects these positions according to the displacements of each joint and the $k$ means clustering. The simulation shows that the selected solutions for the manipulation are all in the finger work space. It proves that this model can improve the manipulability for dual-hand robots. Also, the manipulation experiments show that the robot can avoid the kinematic singularity when approaching the desired object position in bimanual manipulation.

## 1. Introduction

How to determine a manipulation strategy to approach the desired position is significant. It means the kinematic uncertainty between grasped object and robotic fingers. Usually, people can solve it by path planning, error tracking and inverse kinematics. For the bimanual manipulation, the inverse kinematic solutions are diverse and unknown [1]. It affects the dexterity of the manipulation.

The inverse kinematics of the robotic finger could have some missing solutions, multiple solutions or have not solutions [2]. They meant the fingertip can not touch the object surface for the in-hand manipulation. Usually, for the multi-finger manipulation framework, researchers will compute the inverse solutions through the fingers Jacobian matrix [3]. It simplifies the process of modelling for the speed or force control of the manipulation. However, the inverse computation can be complex and coupled from an object velocity to each finger velocity. The reference [4][5] emphasize to decompose the manipulation framework. It means researchers can only consider the inverse computation based on the contact point on the fingertip which can always grip the object. Also, the decomposition means the computation can be separated into a map involves object pose and fingertip pose. Here the constraints are important to seek the optimal combination of two inverse solutions.

Usually, the inverse solution which is computed by the Monte Carlo method is non-linear and not unique. Here researchers need to define the random density function or motion cost function for seeking the optimal solution [6]. The random sample can be the joints values or the positions. The function is piece-wise with the weights for avoiding the singular position of each finger joint in the work space [7]. Besides, the clustering or dimension reduction can simplify these random samples from 3D to lower dimension [8].

For the kinematic simulation, the algorithm design for approaching or seeking is necessary [9]. During the manipulation, robotic hands will create a large number of discrete position samples. For the proper kinematic solution, there are 3 methods. First, the algebraic approach for the finger to follow the movement of contact point so that the object can be well grasped. Then, it is the iterative approach for avoiding the singularity of finger in the work space. Third, it is the geometric approach which can be suitable for the fuzzy logic or neural networks [10], [11].

This paper describes the related works of kinematic modelling and bimanual strategies for the dual-hand robots in Sec. 2, the problem statement of robotic fingers computation for the in-hand manipulation in Sec. 3. Besides, it will describe the approach of manipulation kinematics modelling in Sec. 4. After that, the experiments and simulations will be discussed in Sec 5 and the conclusion with the future works will be discussed in Sec. 6.

## 2. Related works
### 2.1. Finger-object model

The finger-object model demonstrates the geometric relations. It can evaluate or estimate the states of the robotic dynamics or the kinematics of a finger-object system [12]. Also, researching are proposing that 2-D or plane finger-object model can solve most of manipulation tasks [13]. They would like to set a virtual plane or the virtual contact points [14] based on kinematics. In their methods, the robot can estimate and guarantee the rolling manipulation without object information. The weakness is that the manipulation stability and the modes or strategies are limited in these methods.The Monte Carlo method provides the kinematic solution by the numerical simulation [15]. Based on that, researchers can obtain a large number of configuration samples for the analysis of the work space. It also guarantees that configuration which is inversely solvable. For the inverse computation of the robotics, these samples can be the foundation of optimizing the solutions and can extend the manipulation strategies.Based on the finger-object model, we can analyze the relation between the object and the work space.

**2.2. Learning for Solution.**

For processing these random positions from the the Monte Carlo method, there are two methods based on the learning techniques [16]. The first is to learn based on the object positions via the PCA (principal component analysis). The second is to learn based on the valid fingers configurations via the k-means clustering. The key is to reduce the dimensions of the random samples. When the robot learned the features from the object path or the finger work space, the robot can be heuristic to seek the analytical solutions. It means the improvement of the manipulability through a heuristic method.

In this paper, based on the k-means clustering, we obtain the random fingertip positions which are near the object positions. After that, the robot can determine the fingertips to approach these positions. It means that the robot can avoid the singularity in the inverse kinematics. Also, it promotes the analytical solutions of the fingertips to be object-centered.

**2.3.In-hand Manipulation.**

In the last decades [17], researchers believed that the valid grasp gestures and dexterity are the foundation of in-hand manipulation. And they tried to transfer the grasp gestures into the manipulation configuration based on the taxonomy of grasp gestures. The valid grasp gestures for the manipulation in the inverse kinematics should be solvable. For the most of manipulation situation, the initial grasp gestures are known. Also, according to the manipulation tasks, the desired object positions can be known via the path planning. However, there are few of gestures for dual-hand manipulation. Also, the inverse kinematic computation is more challengeable. Before inputting the inverse solution into the robotic control or dynamic model, we hope to seek the optimized fingertip positions. It should be effective, efficient and appropriate for manipulation. Under these positions, the robot can constantly receive the incremental parameters for approaching the relative goal pose. Here the success rates and the accuracy of manipulation is important [18].

## 3. Problem statement.

The problem we want to solve is that seeking the inverse analytical solution for robotic fingers so that the robot can move the object bimanually. Here optimized and appropriate joints positions of fingers will be selected according to the relative motion of the object and the contact points. There are 2 metrics for the solutions that will be considered on this research:

   (1) For each contact point, its position should be linearly correlated with the movement of the object position.
   (2) For the robotic fingers, the configurations generated via a random phase in work space can avoid the singularity during the manipulation.

The joints values can be derived in the hand-object plane reference via the geometric relation between the length of finger link and the displacement of finger joint. For the bimanual or multi-finger manipulation, each finger has its own working plane which involves the initial and desired position of the object and finger joints. Here we initially compute the positions of contact point and then it determines the positions of finger joints. Besides, in the plane reference frame, we can consider 2 position elements and 1 orientation element for expressing the possible movement of the object. Before the manipulation, the two hands have well grasped the object. It means the initial configurations for manipulation is known by the robot. In this paper, some typical cases of the bimanual manipulation will be considered.

## 4. Manipulation kinematics modelling.

Initially, the robot receives the object and joints position in the working plane. Then, the strategies will be determine according to the relation between link length of finger and the displacement of joints. Finally, the manipulation can be executed according to the proportion of object relative motion and joints displacements.

### 4.1 Seeking inverse solution of a finger.

When 2-finger well grasped the unknown object, we consider 2 cases:

**Case 1:** The position adjustment of object by the coordinated motion of fingers. For the planar reference frame of the object, the distance is fixed between object position and contact points. This distance is kept by finger configurations.
**Case 2:** The orientation adjustment of object by fingers rolling. Here the contact points are adjusted according to the object orientation and they can be kept by finger configurations.

For case 1, in the working plane of a robotic finger, we need to derive the new positions of finger joints. Here a robotic finger has 3 links, then, exists:

**Step 1:** In Figure 1 (a), compute the new position for the object:
$$P = P(0) + \Delta P \qquad (1)$$
**Step 2:** For keeping the contact points, correspondingly, compute the new position of the contact point, $c$, according to the relative motion of object, $\Delta P$, as:
$$c_1 = c_1(0) + \Delta P \qquad (2)$$
Here $c(0)$ is the initial contact point.

**Step 3:** For seeking the appropriate position of joint $q_1^3$, exists the constrains as:
$$\begin{cases} l_1^3 = ||q_1^3 - c_1||_2 \\ e_1^3 = ||q_1^3 - q_1^3(0)||_2 \end{cases} \qquad (3)$$

Also, for the position of joint $q_1^2$, exists:
$$\begin{cases} l_1^2 = ||q_1^3 - q_1^2||_2 \\ e_1^2 = ||q_1^2 - q_1^2(0)||_2 \end{cases} \qquad (4)$$

As the constraint for seeking the appropriate positions of finger joints, we can set a cost function $f(t)$ which can be expanded in the running time $t$. It represents the cumulative displacement of $e_1^3$ and $e_1^2$ in Figure 1 (a). The displacements should be minimized from the ending joint, and should be equal to the relative motion of unknown object as:
$$f(t)_1 = \int_0^1 \frac{||\Delta P||_2}{e_1^3 + e_1^2 \cdot t} dt \qquad (5)$$

Here is the running time of the manipulation. Then, it is the similar step of setting constraints until the joint on the palm, $q_1^1$. For relation between $q_1^1$ and $q_1^2$, due to the palm of the robotic hand is fixed, exists:
$$\begin{cases} q_1^2 = q_1^2(0) + [l_1^1 \cdot \cos\theta_1^1, l_1^1 \cdot \sin\theta_1^1]^T \\ \theta_1^1 = \arccos(1 - \frac{(e_1^1)^2}{2 \cdot (l_1^1)^2}) \end{cases} \qquad (6)$$

*Table 1*：Position solutions of finger joints

*Input*:
initial parameter $[P(0), c(0), q_1^1(0), q_1^2(0), q_1^3(0)]$; desired motion $\Delta P$;
link length of a finger $[\ l_1^1,\ l_1^2,\ l_1^3]$;
*Output*:
group of finger joints positions $Q_1$;
1: *Randomly generate* $[q_1^1, q_1^2, q_1^3]$;
2: $e_1^3 \leftarrow ||q_1^3 - q_1^3(0)||_2$; $e_1^2 \leftarrow ||q_1^2 - q_1^2(0)||_2$;
3: $f(t) \leftarrow \int_0^1 \left(\frac{||\Delta P||_2}{e_1^3 + e_1^2 \cdot t}\right) dt$

4: **If** $f(t) \neq 1$
5:    **then** *Select other random samples of* $[q_1^1, q_1^2, q_1^3]$.
6:    **else if** $l_1^3 = ||q_1^3 - c_1||_2$ && $l_1^2 = ||q_1^3 - q_1^2||_2$ && $l_1^1 = ||q_1^2 - q_1^1||_2$
7:       **then** $Q_1 = [q_1^1, q_1^2, q_1^3]$
8:    **else** *Select other random samples of* $[q_1^1, q_1^2, q_1^3]$.

**Step 4:** Based on the Monte Carlo method, generate the random position samples for each joint. Then, filter these samples according to Table 1.

**Step 5:** Based on these finger joints positions, we can derive eventually derive the joint value, $\theta_i^j$, for the manipulation. Each joint value can be derived by:

$$\theta_i^j = \arccos(\frac{\|q_i^{j+1}(0)-q_i^j\|_2^2-(e_i^j)^2+(l_i^j)^2}{2\cdot\|q_i^{j+1}(0)-q_i^j\|_2\cdot l_i^j}) \tag{7}$$

Here $i$ is the number of finger for the multi-finger manipulation. $j$ is the number of joints in a robotic finger. Besides, for the ending joint in Fig. 1 (b), $q_1^3$, it can be denoted as:

$$\theta_1^3 = \arccos(\frac{\|c_i(0)-q_1^3\|_2^2-(e_1^3)^2+(l_1^3)^2}{2\cdot\|c_i(0)-q_1^3\|_2\cdot l_1^3}) \tag{8}$$

For case 2, compared with the case 1, the difference is that we need to derive the new positions of the contact points according to the angle of planar orientation, $\varphi$. In Figure 1(b), exists:

$$c_1 = c_1(0) + \|P - c_1(0)\|_2\cdot[\cos\varphi, \sin\varphi]^T \tag{9}$$

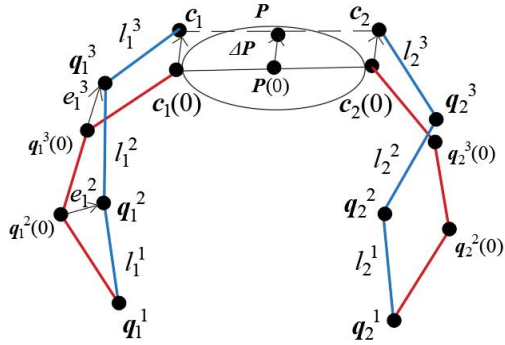

(a) Coordinated manipulation for linear object motion

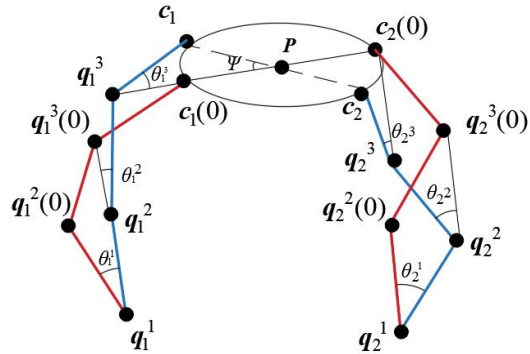

(b) Object rolling by positions change of fingers

**Fig. 1 Planar manipulation cases**

In Figure 1, the red lines meant the initial configurations of robotic fingers. The blue lines meant the appropriate configurations for coordinated or rolling manipulation. It is approachable by adjusting joints values according to those angles $\theta_i^j$. It can be derived by those points which can be demonstrated via 2-dimensial position vectors.

**4.2 Inverse solution of multiple finger.**

Based on 2-finger case in Figure 1, assuming that the linear relation between object motion and joints displacements can be:

$$\gamma \cdot ||\Delta \boldsymbol{P}||_2 = \gamma_1 \cdot f(t)_1 + \gamma_2 \cdot f(t)_2 + \gamma_3 \cdot f(t)_3 \tag{10}$$

Here $\gamma$ and $\gamma_i$ are the weight which can be derived according to the displacements in Eq. (5). They determine the manipulation strategies among fingers. Based on the weight of fingers displacements, we can consider further 2 cases:

**Case 1:** Manipulate unknown object by 3 fingers in a robotic hand.

**Case 2:** Rolling unknown object by 4 fingers from 2 robotic hands. Each hand provides 2 fingers for the simultaneous manipulation.

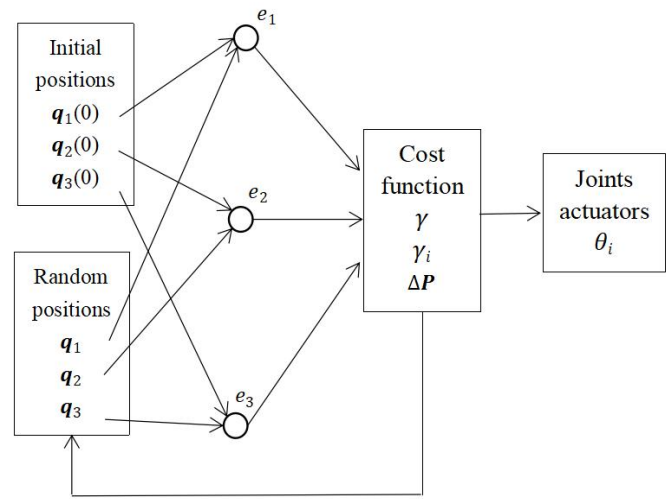

**Fig. 2. 3-finger cases in the work space of the manipulation**

Figure 2 shows the relations between the displacements of each finger joint and the cost function. The displacements can be computed by the initial positions and the random positions. Based on Eq.(10), the cost function determines that whether joints values can be outputted or random positions need to be updated. For bimanual manipulation, the weights in the function determine the strategy the to approach the desired position. Therefore, exists steps:

**Step 1:** Compute the new position of the object according to Equation (1);

**Step 2:** For each finger, randomly generate the joints positions and then filter according to Table 1;

**Step 3:** Compute the displacement of each joint and then derive the appropriate weights according to Equation (10);

**Step 4:** Compute each joint value according to Equation (6), (7) and (8), then, cluster according to the controid which is set based on $P(0)$, $p_i$ and $P$.

**5. Simulation and experiment.**
**5.1 Platform overview.**

Our simulation and experiment tools include:

(1) In the experiment, using ROS-Rviz for controlling the mobile anthropomorphic dual-arm robot (MADAR);

(2) The Matlab/ Simulink and robotic toolbox for the analysis of robotic joint values;

(3) Solidworks/ Motion for the hand-object kinematics;

Here the MADAR in this paper will be used to test the robotic manipulation capability. This dual-arm platform published by Suárez et al. [21] in order to research the related human-like robotics. The other hardware devices as the lower machine of the MADAR include two UR5 robotic arms. With the development of dual-hand manipulation, the grasp capability is extending to 15kg. For the safe human-robot interaction, the MADAR is equipped with a RGB-D camera and the tactile sensors on the robotic fingertips.

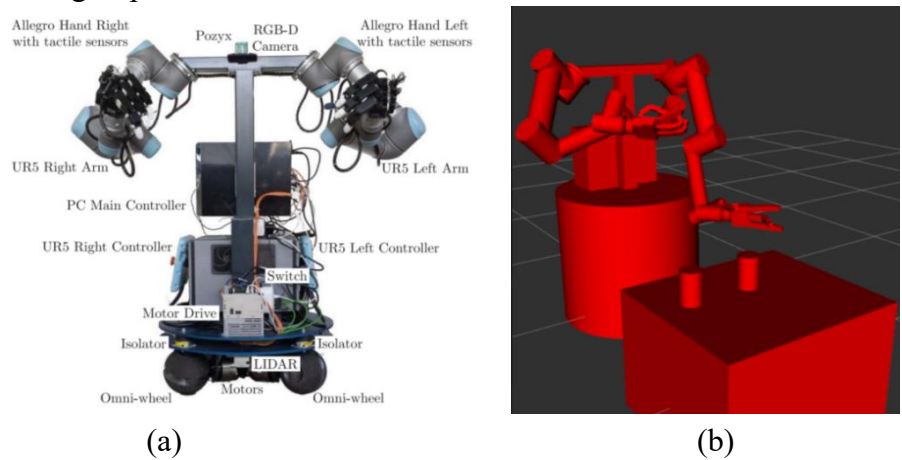

(a) (b)

**Fig. 3. MADAR and its ROS/Rviz model.**

Figure 3 shows the MADAR and the model in ROS/Rviz which can be used to bimanually manipulate. In this research, the simulation and experiment will be demonstrated by 2 cases:

(1) Simultaneous manipulation for rolling the object to adjust its orientation.

(2) Simultaneous manipulation for coordinately moving the object to adjust its position in a table.

We will use 3 ending joints of each UR5 arm as the wrist of the robot and 2 Allergo hands with the tactile sensors on the fingertips to research the bimanual manipulation. Besides, the RGB-D camera is necessary for this paper to calibrate the objects positions and fingers positions. The simulation will verify the manipulability based on the desired angle of the fingertip. After that, the simultaneous and interleaved features will be analyzed based on our algorithm.

**5.2 Tests on the 2-finger manipulations**

In the work space, we use the MADAR to grasp 4 types of objects by its two hands. The object include sphere, cylinder, cone and cube. Using 2 fingers in a robotic hand, we can generate 4 clusters to evaluate the same motion path of the object as below:

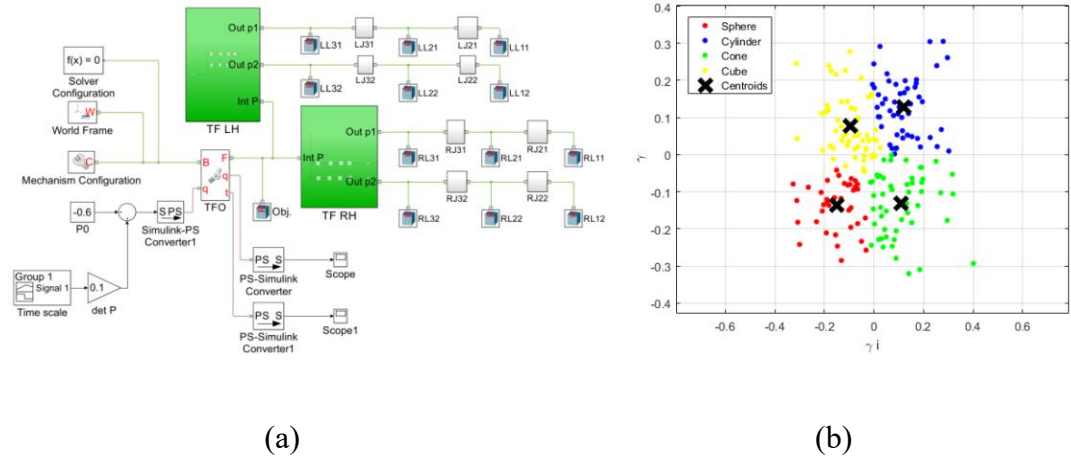

(a)                    (b)

**Fig. 4. K means clustering for 2-finger manipulation.**

Figure 4 (a) shows the scheme for 2-finger case in Matlab/ Simulink. Here we can set the desired object position via a joint module of the Simulink. The shape parameters of unknown objects can be set in the Obj. Module. The scope module can obtain the displacement date of the object. The module of TF LH and TF RH can determine the angles and configurations for the B2F based on our proposed approach. Based on that, Figure 4 (b) shows the weights distributions for manipulating 4 different objects. Here $\gamma_i$ is counted as the average value of the fingers weights. The objects include the sphere, cylinder and ellipse. Each object respectively is held by the Allegro hand with different grasp configuration. At the same time, the prototype of the robotic hands are establish in the simulations. Here the contact model is set as the soft-finger contact with the friction coefficient μ = 0.6. After filtering in Table 1, there are 50 configurations for each object and we clustered the 4 centroids of their weights. Based on the centroids, the proper weights can be derived and it prevented falls for 20-degree rolling motion of the objects.

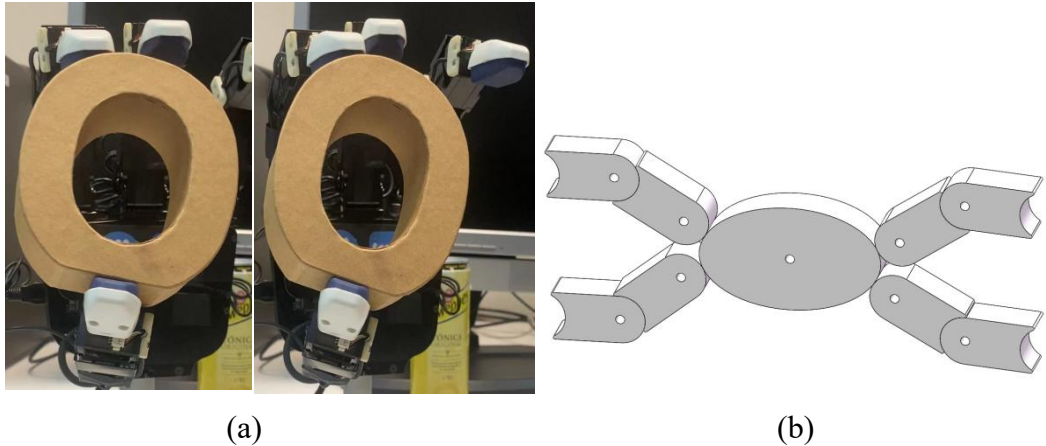

(a)                          (b)

**Fig. 5. Manipulation of 2 and 4-finger cases.**

Figure 5 shows snapshots of the manipulation for the ellipse object. We manually set initial fingers configurations for well grasping the object. Then, the manipulation starts and the hands moves to an appropriate configuration. For this 2-finger manipulation, in Figure 5 (a), we set the desired object orientation so that the object can be rolled around its centroid for 15 degrees. In Figure 5 (b), for the similar 15-degree movement in Solidworks/Motion, 2 hands will respectively use 2 fingers. Based on the algorithm of Table 1, the valid manipulation can be executed.

**5.3 Tests for different manipulation cases**

Apart from 2-finger case for the manipulation by single or dual hand. We also analyze the 3-finger manipulation for further verifying our purpose approaching.

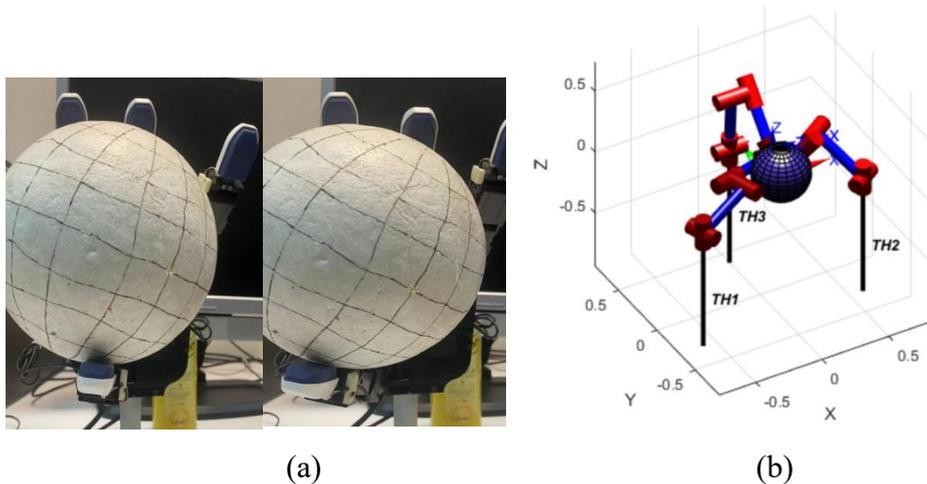

(a)                          (b)

**Fig. 6. 3-finger manipulation in the experiments and simulations.**

Figure 6 shows 3-finger manipulation for a sphere object. Figure 6 (a) is the manipulation experiment by Allegro hand. Figure 6 (b) is the simulation based on 3-finger prototype in Matlab/ Robotic toolbox. Based on the algorithm of Table 1 and the Equation (10), the valid manipulation for the 15-degree rotation of spherical object can be executed.

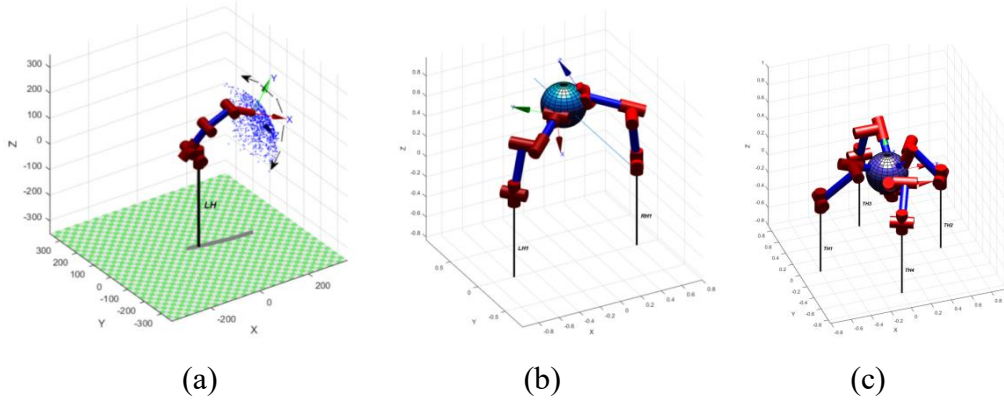

| (a) | (b) | (c) |

**Fig. 7. The simulation for the bimanual manipulation.**

Figure 7 (a) shows the simulation of a robotic finger generated over 100 configuration via the Monte Carlo method which can determine the rolling motion for a spherical object. Figure 7 (b) shows the simulation of 2-finger case in Matlab for executing a linear motion and Figure 7 (c) shows the 4-finger case. The data from these cases are collected and prove that the kinematic singularity can be avoided in the random process of Table 1.

**Tab. 2.** *Error and success rate of the manipulation*

|  | Ellipse | | Sphere | | Cylinder | | Cone | | Cube | |
|---|---|---|---|---|---|---|---|---|---|---|
|  | $e$ | SR | $e$ | SR | $e$ | SR | $e$ | SR | $e$ | SR |
| 2F | 4% | 84% | 4% | 82% | 7% | 76% | 10.3% | 66% | 12% | 64% |
| 3F | 8% | 83% | 5% | 86% | 6% | 74% | 9% | 81% | 13% | 69% |
| B2F | 5% | 85% | 6% | 80% | 8% | 82% | 12% | 77% | 14% | 75% |

**Tab. 3.** *Position and orientation settings for objects*

|  | Ellipse | | Sphere | | Cylinder | | Cone | | Cube | |
|---|---|---|---|---|---|---|---|---|---|---|
|  | *Ini.* | *Des.* | *Ini.* | *Des.* | *Ini.* | *Des.* | *Ini.* | *Des.* | *Ini.* | *Des.* |
| $x$ | 35 | 35 | 32 | 32 | 36 | 36 | 47 | 47 | 49 | 49 |
| $y$ | 28 | 28 | 26 | 26 | 30 | 30 | 22 | 22 | 32 | 32 |
| $z$ | 45 | 45 | 47 | 47 | 46 | 46 | 41 | 41 | 44 | 44 |
| $\rho$ | 37 | 43 | 36 | 39 | 35 | 37 | 31 | 37 | 36 | 42 |
| $\beta$ | 62 | 74 | 58 | 67 | 66 | 72 | 56 | 62 | 59 | 66 |
| $\gamma$ | 25 | 29 | 25 | 29 | 28 | 32 | 29 | 35 | 25 | 31 |

Table 2 shows the relative errors *e* and the success rates (SR) of separately manipulating 5 objectives in Matlab. For the work space, the 6-dimensional initial setting (*Ini.*) and the desired positions of the object (*Des.*) in the simulation are shown in Table 3. The errors are counted based on the desired object position and the actual object position in the simulation software. The success rates are counted according to the falls in the 50-time same manipulation. We can see that our proposed approach can control the relative errors which are less than 15%. For the success rates, the 2-finger manipulation by a robotic hand (2F) has highest SR when rolling the ellipse object. The 3-finger manipulation by a robotic hand (3F) has highest SR when rolling the sphere object. The 2-finger manipulation by 2 robotic hands (B2F) has SR which are all above 75%.

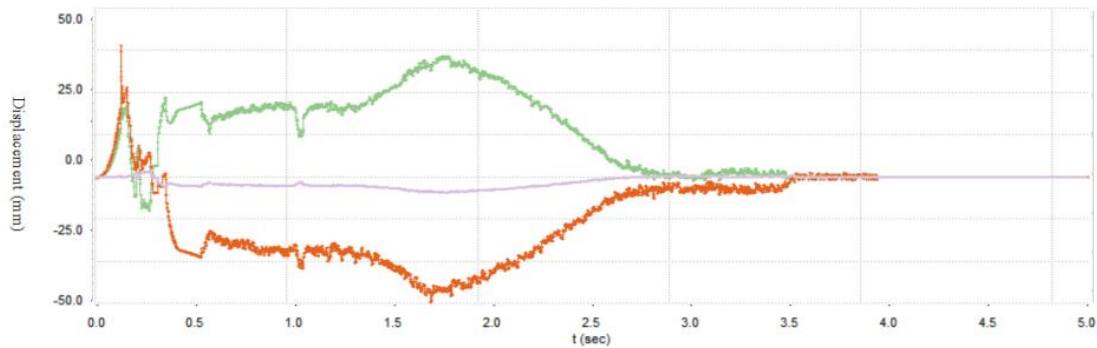

(a)

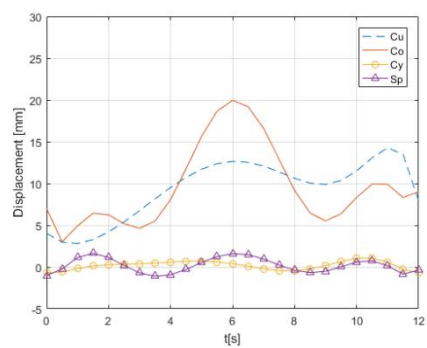

(b)

**Fig. 8. Displacement tracking for the manipulation simulation.**

Figure 8(a) shows displacement tracking of 2F manipulation through the solidworks/ motion. Here the green line means the contact point of left finger, the pink line means the ellipse object and the red line means the contact point of right finger. Two fingers in the path of displacement shows both the symmetry and similarity. The object shows the vibration of the object due to the simulation setting of the contact mode and the gravity. Apart from that, Figure 8(b) shows displacement tracking of B2F manipulation through the through the Matlab/ Simulink. Here the we use the cube (Cu), cone (Co), cylinder (Cy) and sphere (Sp) object. The simulation shows that for the same manipulation to rotate the different object for 15 degrees in 12 seconds according to the Table 3. The displacement of manipulating cone is the biggest, and the cylinder and the sphere are smallest.

## 6. Conclusion.

This paper presented an approach for bimanually manipulating unknown object. It demonstrated a geometric reasoning process based on planar kinematics. The experiments completed based on 2-finger and 3-finger manipulation cases, and the 4-finger manipulation by 2 hands. Besides, 4 types of the objects have tested by the 2 hands. Here each hand provided 2 robotic fingers. For the simulation and experimental results, it showed that our approach can provide valid solutions for adjusting the position and orientation of unknown object. In the working plane, our approach has the manipulability. It showed in that solution can prevent the objects falls and it can provide the optimized outputs for the relative object movement.

Also, we found the limitations for the bimanual manipulation that is the accuracy influence from the Monte Carlo method. It meant we need to further seek the appropriate random samples based on different features of the unknown objects. Besides, future works include this accuracy issue for more manipulation cases. Besides, the optimization based on Monte Carlo method is worth to be discussed.